%% file: main.tex
\documentclass{article}

\usepackage[preprint]{neurips_2022}

\input{tex/packages.tex}

\title{On the Planning, Search, and Memorization Capabilities of Large Language Models}

\author{%
  Yunhao Yang\\
  Department of Computer Science\\
  University of Texas at Austin\\
  Austin, TX 78705 \\
  \texttt{yunhaoyang234@utexas.edu} \\
  \And
  Anshul Tomar\\
  Department of Computer Science\\
  University of Texas at Austin\\
  Austin, TX 78705 \\
  \texttt{anshulmanas@gmail.com} \\
}

\begin{document}

\maketitle

\input{tex/00-abstract.tex}
\input{tex/01-introduction.tex}
\input{tex/02-background}
\input{tex/03-planning-domain}

\input{tex/04-graph-search}
\input{tex/05-adversarial-planning}
\input{tex/06-finetuning}

\input{tex/07-conclusion}

\bibliographystyle{plainnat}
\bibliography{ref}

\input{tex/A-appendix}

\end{document}

%% file: tex/packages.tex
\usepackage[utf8]{inputenc} 
\usepackage[T1]{fontenc}    
\usepackage{hyperref}       
\usepackage{url}            
\usepackage{booktabs}       
\usepackage{amsfonts}       
\usepackage{nicefrac}       
\usepackage{microtype}      
\usepackage{xcolor}         

\usepackage{graphicx}
\usepackage{doi}
\usepackage{latexsym}
\usepackage{amsmath}
\usepackage{amssymb}
\usepackage{multirow}
\usepackage{pgfplots}
\usepackage{csquotes}
\pgfplotsset{compat=1.18}
\usetikzlibrary[pgfplots.groupplots]

\usepackage[ruled]{algorithm2e}
\SetKwComment{Comment}{/* }{*/ }
\SetKwInOut{Parameter}{parameter}
\usepackage{array}
\usepackage{makecell}

\usepackage{newfloat}
\usepackage{listings}
\usepackage{xcolor}
\usepackage{amsthm}
\usepackage[english]{babel}
\theoremstyle{definition}

\usepackage{macros}

\usepackage{tikz}
\usetikzlibrary{arrows, automata, positioning}
\tikzset{auto, >=stealth}
\tikzset{every edge/.append style={shorten >= 1pt}}
\tikzset{
    main node/.style={circle,draw,minimum size=1cm,inner sep=0pt},
}

\lstdefinelanguage{PDDL}
{
  sensitive=false,    
  morecomment=[l]{;}, 
  alsoletter={:,-},   
  morekeywords={
    define,domain,problem,not,and,or,when,forall,exists,either,
    :domain,:requirements,:types,:objects,:constants,
    :predicates,:action,:parameters,:precondition,:effect,
    :fluents,:primary-effect,:side-effect,:init,:goal,
    :strips,:adl,:equality,:typing,:conditional-effects,
    :negative-preconditions,:disjunctive-preconditions,
    :existential-preconditions,:universal-preconditions,:quantified-preconditions,
    :functions,assign,increase,decrease,scale-up,scale-down,
    :metric,minimize,maximize,
    :durative-actions,:duration-inequalities,:continuous-effects,
    :durative-action,:duration,:condition
  }
}

%% file: tex/00-abstract.tex
\begin{abstract}
  The rapid advancement of large language models, such as the Generative Pre-trained Transformer (GPT) series, has had significant implications across various disciplines. In this study, we investigate the potential of the state-of-the-art large language model (GPT-4) for planning tasks. We explore its effectiveness in multiple planning subfields, highlighting both its strengths and limitations. Through a comprehensive examination, we identify areas where large language models excel in solving planning problems and reveal the constraints that limit their applicability. Our empirical analysis focuses on GPT-4's performance in planning domain extraction, graph search path planning, and adversarial planning. We then propose a way of fine-tuning a domain-specific large language model to improve its Chain of Thought (CoT) capabilities for the above-mentioned tasks. The results provide valuable insights into the potential applications of large language models in the planning domain and pave the way for future research to overcome their limitations and expand their capabilities.
\end{abstract}

%% file: tex/01-introduction.tex
\section{Introduction}

The fast growth of large language models, such as the Generative Pretrained Transformer (GPT) series, significantly impacts various disciplines, from natural language processing and artificial intelligence to healthcare \citep{chintagunta2021medically, nori2023capabilities, thirunavukarasu2023large}, finance \citep{leippold2023thus, finance2}, and beyond \citep{DBLP:conf/emnlp/ShinZBPPSM20}. These models have revolutionized tasks such as machine translation, sentiment analysis, text summarization, and question-answering, enhancing human-computer interactions and enabling more efficient and accurate information retrieval. In addition, the vast amounts of data these models are trained on allow them to generate human-like responses and perform tasks that were once considered exclusive to human intelligence.

We examine the capability of the current state-of-the-art language model---GPT-4---on planning and search \citep{gpt-4}. Despite its impressive performance in natural language processing tasks and its ability to generate human-like text, GPT-4 is not explicitly designed for executing planning or search algorithms. However, it can provide valuable insights and guidance on various planning and search techniques and domain-specific knowledge for constructing heuristics or evaluating different approaches. GPT-4's vast knowledge base allows users to ask questions and explore diverse aspects of planning and search. 

We indicate the fields in planning that can be solved by large language models and the limitations of language models. The introduction of large language models significantly impacts many fields, such as natural language processing; hence we want to examine its impact on the field of planning.

Existing works \citep{valmeekam2022large, huang2022inner, singh2023progprompt, LinHLGS023} demonstrating the capability of language models on planning are heavily focused on plan generation but lack the exploration of path search, memorization in planning, and planning in adversarial settings.

We provide a comprehensive examination of the capability of GPT-4 in the field of planning and indicate its limitations for future research. Additionally, we attempt to improve the performance of an LLM by fine-tuning it on tasks like planning domain, graph search, and adversarial search to see if we are able to improve the predictions of these models for these aforementioned tasks. These models are capable of addressing various planning tasks, such as providing general information on planning algorithms, generating heuristics, and discussing different planning techniques. However, they are not specifically designed to perform planning tasks directly, as their primary function is to understand and generate text. Language models have limitations in handling real-time interactive scenarios and lack the ability to learn and adapt beyond their training data. Consequently, while large language models can provide valuable insights and guidance in the realm of planning, their utility is constrained by these limitations, and they cannot fully replace specialized planning algorithms or tools designed to address specific planning problems.

We provide an empirical analysis of how GPT-4 performs on planning domain extraction, graph search path planning, and adversarial planning.
We found that GPT-4 is effective at extracting key components of planning domains from textual descriptions, allowing for the generation of structured representations suitable for use in automated planning systems. In graph search, GPT-4 exhibits the capability to understand the searching algorithm and find an optimal path based on the algorithm. However, such capability is limited once the graph becomes complicated. Moreover, we show its capability to generate heuristics for adversarial planning and its limitation in performing adversarial search algorithms. The lack of memorization during planning is a main factor that limits the large language model to planning in adversarial settings.

%% file: tex/02-background.tex
\section{Related Work}

Several works have used large language models for zero-shot planning; however, their planning either assumes the planning domain is acquired, or the outcomes are static. Some works \citep{glm2fsa, saycan} only generate static outcomes, while LLM-Planner \citep{llm-planner} and LM-Nav \citep{llm-nav} require prior knowledge of specific fields to define the planning domain.
Existing works \citep{embodied, glm2fsa} have demonstrated the capability of these models. Large language models are sources with a wide range of knowledge, including domain-specific knowledge. However, existing works have not dived into the planning and searching capabilities of these models, especially in complex problem or adversarial settings.

In this work, we explore the capabilities of large language models on planning domain generation, graph search, planning state memorization, and adversarial planning. The work reveals some limitations of large language models, which lead to potential future directions for improving these models.

\section{Preliminaries}
\paragraph{Large Language Models (LLMs).}
LLMs are machine learning models designed to process and understand natural language, such as human speech and text. These models are typically large-scale neural networks, trained using massive amounts of data, often on the scale of billions of words or more, to learn patterns and structures in language.

LLMs are capable of a wide range of natural language processing tasks, such as language translation, sentiment analysis, text classification, and speech recognition. They can generate new text based on the input prompt they received and create original content such as news articles, essays, or even poetry.

One example of a large language model is OpenAI's GPT (short for ``Generative Pre-trained Transformer") series \citep{brown2020GPT3, gpt-4}, with GPT-4 being the most current iteration. Compared to existing LLMs, GPT-4 is also able to understand image inputs and perform better on logic reasoning. These models have demonstrated remarkable performance across a wide range of NLP tasks, revolutionizing the field of AI and enabling new applications in various domains.

\paragraph{Planning Domain.}
A planning domain refers to a formal description of a specific problem space or environment \citep{pddl}. It consists of the rules, constraints, and actions that define the structure of the problem and the ways in which it can be solved. The goal of automated planning is to find a sequence of actions that can transform the initial state of the domain into a desired goal state.

A planning domain generally consists of the following components:
\begin{itemize}
    \item Objects: The entities or items that exist within the domain, such as people, locations, or resources.
    \item States: A description of the various conditions or configurations of the objects in the domain.
    \item Actions: The operations or steps that can be taken to modify the state of the domain. Actions usually have preconditions that must be satisfied before they can be executed and effects that describe how the state changes when the action is performed.
    \item Initial state: The starting configuration of the domain from which the planning process begins.
    \item Goal state: The desired configuration or set of conditions that the planning process aims to achieve.
\end{itemize}

Planning Domain Definition Language (PDDL) is a formal language used to describe planning problems and domains in the field of automated planning. PDDL separates the description of a planning problem into two parts: the domain and the problem. The domain defines the general structure of the problem, including the available actions and their effects, while the problem specifies the initial state and the goal state for a particular instance of the problem.

In addition to the components of general planning domains, PDDL consists of a set of predicates, which is a set of properties or relations that describe the state of the objects in the domain.

\paragraph{Graph Search.}
Graph search is a type of algorithm used to explore and navigate graphs, which are mathematical structures consisting of nodes (also called vertices) connected by edges. In graph search, the algorithm starts at a given node and systematically explores the graph by visiting its neighboring nodes in a specific order until it reaches a target node or a goal state.

The goal of graph search is to find the shortest or most efficient path between two nodes in a graph. There are several different types of graph search algorithms, including:
\begin{itemize}
    \item Breadth-first search (BFS): This algorithm explores all the neighbors of a node before moving on to the next level of nodes. BFS is guaranteed to find the shortest path between two nodes in an unweighted graph.
    \item Depth-first search (DFS): This algorithm explores one branch of the graph as far as possible before backtracking and exploring another branch. DFS can be used to find all paths between two nodes in a graph, but it may not find the shortest path.
    \item Dijkstra's algorithm: This algorithm is used to find the shortest path between two nodes in a weighted graph. It works by assigning a tentative distance to each node and updating the distance as it explores the graph.
    \item A* search: This algorithm is similar to Dijkstra's algorithm but uses a heuristic function to guide the search toward the goal node. A* search is often used in pathfinding in video games.
\end{itemize}
Graph search algorithms can be used to solve a wide range of problems, but the choice of algorithm depends on the specific problem and the characteristics of the graph being searched.

\paragraph{Adversarial Planning.}
Adversarial planning is a type of planning problem where the planner is required to generate a plan that can anticipate and react to the actions of an adversarial agent. In this type of problem, the planner must take into account the actions of the adversary and try to find a plan that maximizes the chances of success while minimizing the impact of the adversary's actions.

Adversarial planning is commonly used in game theory, where it is used to model the strategies and actions of two or more players engaged in a game. In this context, the planner must anticipate the actions of the opponent and develop a strategy that maximizes the chances of winning.

There are several approaches to adversarial planning, including mini-max, in which the goal of the agents is to maximize their own rewards or utility while minimizing the rewards or utility of their opponents, and Monte Carlo Tree Search (MCTS), which uses a search algorithm to simulate the possible outcomes of the planner's actions and the adversary's responses.

Adversarial planning is a challenging problem because it requires the planner to consider not only their own objectives but also the objectives and capabilities of the adversary agents. As a result, it often involves complex decision-making and requires sophisticated algorithms and techniques.

%% file: tex/03-planning-domain.tex
\section{Planning Domain Generation}

In this section, we formulate an approach to generating planning domains using the large language model.
To generate the planning domain, we apply the following procedure: querying a brief description of a task to GPT and transforming the responses into PDDL. If we have prior task knowledge, we can send it to GPT and ask it to generate PDDL from the knowledge. This approach enables the task designers to obtain task knowledge in a formal representation, regardless of the prior information the task designers have. Moreover, after generating the planning domains, the task designer can obtain a task plan by performing a simple path search. We also show the capability of GPT-4 on path search in the later sections.

We start the experiment with a daily-life task---cross the road---to examine the planning domain generation ability. We send the following input prompt to GPT-4:
\begin{lstlisting}[language=completion]
    <prompt> Define a problem and actions for a task "cross the road at traffic light" in PDDL.</prompt>
\end{lstlisting}
The PDDL outputs generated by GPT-4 is presented in Listing \ref{lst:cross}.
\begin{lstlisting}[float=!htb, caption={Define a problem and actions for a task "cross the road at traffic light" in PDDL}, label={lst:cross}, language=PDDL]
(define (domain cross-road)
  (:requirements :strips)
  (:types side)
  (:predicates
    (pedestrian-at ?s - side)
    (traffic-light ?s - side)
    (safe-to-cross ?s - side)
  )
  (:action wait-for-green
    :parameters (?s - side)
    :precondition (and (pedestrian-at ?s) (traffic-light ?s) )
    :effect (safe-to-cross ?s)
  )
  (:action cross
    :parameters (?s1 - side ?s2 - side)
    :precondition (and (pedestrian-at ?s1) (safe-to-cross ?s1) )
    :effect (and
      (not (pedestrian-at ?s1)) (pedestrian-at ?s2) 
      (not (safe-to-cross ?s1))
    )
  )
)
(define (problem cross-road-problem)
  (:domain cross-road)
  (:objects side1 side2 - side)
  (:init
    (pedestrian-at side1)
    (traffic-light side1)
    (traffic-light side2)
  )
  (:goal (and (pedestrian-at side2)))
)
\end{lstlisting}

Then, we query GPT-4 to find a plan by searching through the planning domain:
\begin{lstlisting}[language=completion]
    <prompt> Solve the problem for "cross the road at the traffic light" in PDDL with the actions defined.</prompt>
    <completion> 1. wait-for-green side1</completion>
    <completion> 2. cross side1 side2 </completion>
\end{lstlisting}
The result indicates that the generated planning domain is self-contained, and we can obtain a formal representation of the plan, which solves a zero-shot planning problem.

In addition to daily-life tasks, we can ask GPT-4 to generate the planning domain for some well-known games, such as Tic-Tac-Toe in Listing \ref{lst: tic-tac-toe} and chess in Listing \ref{lst: chess-problem}. However, once the complexity of the game increases, the success rate of generating self-contained planning domains from GPT-4 decreases. A failure example is the chess game in Listing \ref{lst: tic-tac-toe}, which defines the wrong goal state. 
\begin{lstlisting}[float=!htb, caption={Define a problem and a set of actions for tic-tac-toe in PDDL.}, label={lst: tic-tac-toe}, language=PDDL]
(define (domain tic-tac-toe)
  (:requirements :strips :typing)
  (:types player cell )
  (:predicates
    (cell-empty ?c - cell)
    (cell-filled ?c - cell ?p - player)
    (current-player ?p - player)
  )
  (:action place-mark
    :parameters (?p - player ?c - cell)
    :precondition (and
      (cell-empty ?c) (current-player ?p)
    )
    :effect (and
      (not (cell-empty ?c)) (cell-filled ?c ?p)
      (forall (?other - player) 
      (when (not (= ?other ?p)) (current-player ?other)))
    )
  )
)

(define (problem tic-tac-toe-problem)
  (:domain tic-tac-toe)
  (:objects
    x o - player
    c1 c2 c3 c4 c5 c6 c7 c8 c9 - cell
  )
  (:init
    (cell-empty c1) ... (cell-empty c9)
    (current-player x)
  )
  (:goal (or
    ; Check for horizontal wins
    (and (cell-filled c1 x) (cell-filled c2 x) (cell-filled c3 x))
    ...
    ; Check for vertical wins
    (and (cell-filled c1 x) (cell-filled c4 x) (cell-filled c7 x))
    ...
    ; Check for diagonal wins
    (and (cell-filled c1 x) (cell-filled c5 x) (cell-filled c9 x))
    (and (cell-filled c3 x) (cell-filled c5 x) (cell-filled c7 x))
  ))
)

\end{lstlisting}

\begin{lstlisting}[float=!htb, caption={Define a problem and a set of actions for the chess game in PDDL.}, label={lst: chess-problem}, language=PDDL]

(define (problem chess-problem)
  (:domain chess)
  (:objects
    wp1 wp2 wp3 ...... bp6 bp7 bp8 - piece
    a1 a2 a3 ...... h6 h7 h8 - square
  )
  (:init
    (at wp1 a2) (at wp2 b2) (at wp3 c2) ......
    (clear a3) (clear a4) (clear a5) ......
  )
  (:goal (and (at wp1 a7) ))
)

\end{lstlisting}

For more empirical results, we select 100 tasks with different complexities. The tasks are ranged from board games to daily tasks to domain-specific tasks. Then, we query GPT-4 to generate planning domains for those tasks and check the correctness of the generated domains. We show the results in Table \ref{tab: domain}. 
As we can see, GPT-4 can always generate self-contained planning domains but occasionally generate planning domains that do not match human knowledge.

\begin{table}
    \centering
    \caption{Results on Planning Domain Generation using GPT-4. A correct plan means the planning domain is self-contained and matches human knowledge. A wrong plan means the planning domain is self-contained but does not match human knowledge (e.g., chess game in Listing \ref{lst: chess-problem}). A failed plan means the planning domain is not self-contained due to the inconsistency of predicates.}
    \begin{tabular}{||c|c|c|c||}
        \hline
        Total Tasks &  Correct Plan & Wrong Plan & Failed\\
        \hline
        100 & 73 & 27 & 0\\
        \hline
    \end{tabular}
    \label{tab: domain}
\end{table}

\begin{table}
    \centering
    \caption{Results on path search using GPT-4. Length indicates the length of the plan generated by \emph{fast-downward} using the planning domains from GPT-4. The number of tasks indicates how many tasks can be completed in this range of steps. A plan is considered \emph{correct} if the plan generated by GPT-4 is identical to the plan from \emph{fast-downward}.}
    \begin{tabular}{||c|c|c|c|c||}
        \hline
        Length &  $\leq$ 3 & 4---6 & 6---8 & $\geq$ 8\\
        \hline
        Number of Tasks & 68 & 10 & 7 & 15\\
        Number of Correct Plans & 68 & 8 & 4 & 3\\
        \hline
    \end{tabular}
    \label{tab: plan}
\end{table}

Additionally, we further query GPT-4 to solve the planning problem given those generated domains. Since all the planning domains are self-contained, we also run \emph{fast-downward} planner to find a plan and compare it with the plan generated from GPT-4.
The results in Table \ref{tab: plan} indicate that GPT-4 can find plans for simple tasks, but once the task requires more steps, GPT-4 may generate plans with missing disordered actions.

In conclusion, large language models like GPT-4 are useful in planning domain generation. Due to the rich knowledge encoded in these models, we can use them as a knowledge source, with a notice that they are not always reliable for complex tasks. In the path search aspect, GPT-4 can solve very simple path search problems from given planning domains. However, there is no significant advantage to using GPT-4 compared to using a traditional planner.

%% file: tex/04-graph-search.tex
\section{Graph Search}

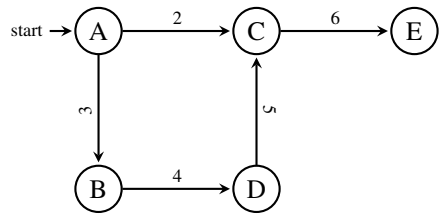
\begin{figure}
    \centering
    \input{figures/graph}
    \caption{A randomly generated direct weighted graph for examining the graph search capability of large language models.}
    \label{fig: graph}
\end{figure}

\begin{figure}
    \centering
    \includegraphics[width=0.5\linewidth]{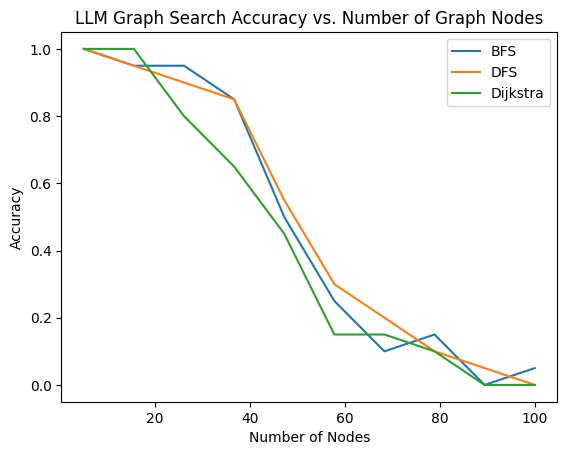}
    \caption{The accuracy of graph searching results generated by GPT-4 on graphs with different nodes.}
    \label{fig: gs}
\end{figure}

In this section, we examine the capability of large language models, such as GPT-4, on graph searches. The examination consists of two aspects: first, whether GPT-4 understands the well-known graph search algorithms, and second, whether GPT-4 can follow the algorithms to find the desired path. Note that we compare the outputs of GPT-4 to the outputs of the graph search algorithms. The outputs are not necessarily the optimal path. We consider GPT-4 to be accurate as long as it can generate paths following the graph search algorithms.

We collect direct weighted graphs from 5 nodes to 95 nodes with a gap of 10 (5, 15, 25, ..., 85, 95 nodes). For each number of nodes, we collect 20 different graphs. An example is presented in Figure \ref{fig: graph}. For each graph, we query GPT-4 to perform three graph search algorithms, depth-first search, breath-first search, and Dijkstra's algorithm, respectively, to generate paths. For the example in Figure \ref{fig: graph}, we query:
\begin{lstlisting}[language=completion]
    <prompt> Run BFS to find the shortest path from A to E.</prompt>
    <completion> Here's the BFS traversal on the directed graph:
    1. Start at node A. Enqueue A and mark it as visited.
    2. Dequeue A. Enqueue its unvisited neighbors, B and C, and mark them as visited.
    3. Dequeue B. Enqueue its unvisited neighbor, D, and mark it as visited.
    4. Dequeue C. Enqueue its unvisited neighbor, E, and mark it as visited.
    5. Dequeue D. There are no unvisited neighbors.
    6. Dequeue E. There are no unvisited neighbors. 
    The BFS traversal visits nodes in the following order: A -> B -> C -> D -> E.
    By using the predecessors, we can reconstruct the shortest path from A to E:
    1. Start at node E.
    2. The predecessor of E is C. So, we have E <-- C.
    3. The predecessor of C is A. So, we have C <-- A.
    Putting it all together, we get the shortest path from A to E: A --> C --> E.</completion>

    <prompt> Run Dijkstra's algorithm to find the shortest path from A to E.</prompt>
    <completion>
    Here's the DFS traversal on the directed graph:
    ......
    The DFS traversal visits nodes in the following order: A -> B -> D -> C -> E.
    Putting it all together, we get a path from A to E: A --> C --> E.</completion>

    <prompt> Run DFS to find the shortest path from A to E.</prompt>
    <completion>
    Here's how Dijkstra's algorithm proceeds:
    1. Distances: A = 0, B = 999, C = 999, D = 999, E = 999. Unvisited: A, B, C, D, E
    2. Visit node A. Update B's distance and set its predecessor as A. Update C's distance and set its predecessor as A. Distances: A = 0, B = 3, C = 2, D = 999, E = 999. Unvisited: B, C, D, E
    3. Visit node C (smallest unvisited distance). Update E's distance and set its predecessor as C. Distances: A = 0, B = 3, C = 2, D = 999, E = 8. Unvisited: B, D, E
    4. Visit node B (smallest unvisited distance). Update D's distance and set its predecessor as B. Distances: A = 0, B = 3, C = 2, D = 7, E = 8. Unvisited: D, E
    5. Visit node D (smallest unvisited distance). No updates. Unvisited: E
    6. Visit node E (smallest unvisited distance). No unvisited neighbors.

    Putting it all together, we get a path from A to E: A --> C --> E.</completion>
\end{lstlisting}

As we indicate in this example, GPT-4 is sufficient to generate accurate paths under all three algorithms. However, as the graphs become more complicated, the accuracy of GPT-4 decreases. In Figure \ref{fig: gs}, we present how the accuracies of GPT-4 on three algorithms decrease as the number of graph nodes increases. Therefore, we conclude that GPT-4 is only capable of simple graph search. But it provides the possibility of decomposing a complex graph into simple graphs and performing graph search.

%% file: figures/graph.tex
\begin{tikzpicture}[
    scale=.7,
    node distance=2.2cm,
    thick,
    every node/.append style={transform shape},
]

\node[state,initial] (A)
    at (3, 6)
    {\Large A};
\node[state] (C)
    at (6, 6)
    {\Large C};
\node[state] (E)
    at (9, 6)
    {\Large E};
\node[state] (B)
    at (3, 3)
    {\Large B};
\node[state] (D)
    at (6, 3)
    {\Large D};

\path[->,sloped]

(A) 
edge[] node[]
    {3}
    (B)
edge[] node[]
    {2}
    (C)

(B) 
edge[] node[]
    {4}
    (D)
    
(C) 
edge[] node[]
    {6}
    (E)

(D) 
edge[] node[]
    {5}
    (C)

;

\end{tikzpicture}

%% file: tex/05-adversarial-planning.tex
\section{Adversarial Planning}

In this section, we explore the capability of the large language model, specifically GPT-4, on adversarial planning. The experiment consists of two components: defining heuristics and applying adversarial search.

First, we choose the simple game Tic-Tac-Toe as an example and query GPT-4 for a proper heuristic:
\begin{lstlisting}[language=completion]
    <prompt> Design a heuristic for adversarial planning in tic-tac-toe.</prompt>
    <completion> Here's a simple heuristic for Tic-Tac-Toe:
    1. Assign a value of +1 for each row, column, or diagonal with only the maximizing player's symbols (and no opponent's symbols).
    2. Assign a value of -1 for each row, column, or diagonal with only the minimizing player's symbols (and no opponent's symbols).
    3. Assign a value of 0 for rows, columns, or diagonals that have both the maximizing and minimizing player's symbols, since these lines are not possible to win.
    4. Sum up the values from steps 1-3 to calculate the total score for the board state.</completion>
\end{lstlisting}

The language model successfully generates a heuristic for adversarial planning in Tic-Tac-Toe. We then manually implement this heuristic and perform the Mini-max algorithm. It turns out that the output of GPT-4 is a workable heuristic.

Second, we examine the performance of GPT-4 on adversarial planning. We query GPT-4 for playing the game with the heuristic defined above:
\begin{lstlisting}[language=completion]
    <prompt> Play tic-tac-toe with me using the Mini-max algorithm with the heuristic defined above.</prompt>
    <completion> Please provide the current state of your Tic-Tac-Toe board, and let me know whether the next move is for 'X' (maximizing player) or 'O' (minimizing player). Then, I'll help you determine the best move according to the heuristic.</completion>
    <prompt>
    Board: ---
           --X
           ---</prompt>
    <completion>Here are the heuristic scores for each possible move:
    O1: O--    O2: -O-    O3: --O    O4: -O-    O5: --O
        --X        --X        --X        -OX        -XO
        ---        ---        ---        ---        ---

    O6: --O    O7: ---    O8: ---    O9: ---
        O-X        O-X        -OX        -XO
        ---        -O-        --O        --O
    O4, O5: 1 (Blocks 'X' from potentially winning in the next move)
    Others: 0 (No immediate advantage or disadvantage)
    According to the heuristic, the best moves for O are O4 and O5.</completion>
\end{lstlisting}
In this example, GPT-4 fails to identify possible moves and memorizes the sequence of previous states, i.e., the existing pieces on the board.
Moreover, even with the misidentified moves, it fails to compute the heuristics. From this observation, we can also claim that GPT-4 is unable to understand the heuristics. Therefore, it has limited capability of graph search algorithms like A* as well.

Due to the failure of adversarial planning in Tic-Tac-Toe, we stopped examining its capability on more complicated tasks and derived our conclusion.

In conclusion, LLMs are capable of generating reasonable heuristics for the adversarial planning of simple games. We examine this capability in the tic-tac-toe example, where the AI-driven heuristic allows for evaluating board positions based on the presence of a player's symbols in rows, columns, and diagonals. LLMs encode rich knowledge and can provide reasonable heuristics for some given tasks.

This simple yet heuristic demonstrates the potential of using AI-generated heuristics to guide and enhance decision-making in various problem domains. 
In conjunction with adversarial search algorithms such as Minimax or Alpha-Beta pruning, these heuristics enable the creation of AI opponents that can effectively compete against human players in simple games like Tic-Tac-Toe.

However, LLMs cannot perform adversarial searching algorithms like Mini-max or Monte-Carlo Tree Search. Due to the fact that GPTs are a series of language models for next-word prediction, they can neither understand the state of the game nor search over all the possibilities (as we addressed the graph search limitation in the last section). Additionally, the language model cannot memorize the sequence of previous states correctly. These factors raise a limitation of LLMs and could potentially be a direction of improvement.

%% file: tex/06-finetuning.tex
\section{Fine-tuning LLM for Logical Reasoning}

Given the subpar performance of LLM on logical reasoning tasks like adversarial planning, we can fine-tune our own language model to check if we could improve its performance on logistic reasoning tasks.

\subsection{Dataset}

The dataset collected comprised three parts which were planning domain generation (7 different tasks), graph search (20 different tasks), and adversarial planning (4 different tasks). For example, for planning domain generation, we queried GPT-4 using seven different problem definitions. Each problem definition generated 10-100 different goal state configurations depending on the problem, resulting in a total of 540 queries. Given each query (only those queries were selected, which we thought would give correct results when passed through LLM), we ran GPT-4 inference on them to get the soft labels for fine-tuning our own LLM. We collected around 1300 queries (appended with the name of the part, e.g., planning domain queries were appended by \texttt{planning domain :} and so on) and soft label pairs across all tasks.

\subsection{Model Selection and Fine-tuning}
To select the correct model to fine-tune, we chose a model small enough that could be easily fine-tuned using the resources available to us and also large enough such that it could infer logically. We chose the Flan T5 base model \citep{flan} released by Google since it meets this criterion. One of the reasons we chose this model was because this model's checkpoints were readily available at HuggingFace, and had a reasonable size of 240M parameters. Also, as shown in \citep{flan}, the model shows the SOTA performance in the CoT dataset \citep{alpaca-cot}, which contains a chain of thought data points like arithmetic reasoning, explanation generation, etc., for the number of parameters it had. For fine-tuning the model, we froze the weights of the original model. We only changed the final layer's weights. Updating the weights of the entire model could have led to catastrophic forgetting, or it would have been fine-tuned properly since our dataset size was too small.

\subsection{Results}

We fine-tuned our LLM model on approximately 1000 data points and evaluated it on the remaining 200 data points. We used the remaining 100 data points as the validation set and used it to stop training the model when we started to see an increase in the validation loss. Given the rapid advances in LLM models, we observed that the outputs of GPT-4 were much better than the Flan models, which were more or less incorrect for every data point we had. In Table \ref{tab: fine-tune}, we compare the outputs of Flan, fine-tuned Flan, and GPT-4 for a specific case of graph search and planning domain generation. As for adversarial search, the flan models are unable to come up with a coherent heuristic, and hence we skip their evaluations. Also, we could not compare the output of the LLM and the ground truth values after planning since all the planning domains provided by the Flan were incorrect. In conclusion, we were able to see minor improvements due to fine-tuning because the original model might have never seen prompts like these but was unable to process the outputs of the fine-tuned models.

\begin{table}[!t]
    \centering
    \begin{tabular}{|l|p{5cm}|p{5cm}|}
    \hline
        & Planning domain & Graph Search  \\
    \hline
       Input query & Q. Define a problem and actions for the task "tic-tac-toe" in PDDL? & Answer the following yes/no question by reasoning step-by-step. Given a graph with vertices A, B, C, D, E where the connections are given as A-B, B-C, C-D, C-E. Run BFS to find the shortest path from A to E. \\ 
       \hline
       Flan & Tic-tac-toe is a word that means "to engage in a classic three-in-a-row game typically played on a grid by two players taking turns to place their respective marks, either X or O, in an attempt to achieve a winning combination." & BFS is a program that calculates the distance between A and E. \\
       \hline
       Flan-finetuned & 
       \begin{lstlisting}
           (define (domain tic-tac-toe)(:requirements :strips :typing :equality))
       \end{lstlisting}
       & BFS is a program that calculates the distance between A and E. Exploring A gives B. Exploring B gives C. Exploring C gives D and E. Therefore, the path is A - B - C - E.\\
       \hline
       GPT-4 &\begin{lstlisting}
           (define (domain tic-tac-toe)(:requirements :strips :typing :equality)
           
           (:types  cell)
           
           (:predicates
                (empty ?c - cell)
                (x ?c - cell)
                (o ?c - cell)
              )
            ....
       \end{lstlisting}& 
       \begin{lstlisting}
   To determine if we can run Breadth-First Search (BFS) to find the shortest path from A to E, let's go through the BFS algorithm step-by-step for the given graph.
   Start at the source vertex A.
   Explore all neighbors of A: (A-B)
   Explore all unvisited neighbors of B: (B-C)
   C has two unvisited neighbors: (C-D) and (C-E)
   At this point, we've reached vertex E, so we can stop the algorithm. The shortest path from A to E is A-B-C-E.
       \end{lstlisting}\\
       \hline
    \end{tabular}
    \caption{The comparisons between the outputs of various models for planning domains and graph search.}
    \label{tab: fine-tune}
\end{table}

%% file: tex/07-conclusion.tex
\section{Conclusion}

Large language models can play critical roles in planning due to their large knowledge domain. The up-to-date large language models encode rich real-world knowledge and can make logical inferences to a certain extent. We provide examples to demonstrate that these models can generate self-consistent planning domains of given tasks without any prior information provided. This capability enables language models to do zero-shot planning. Moreover, the language models can perform graph searches on small-scaled graphs, indicating their great potential in searching. However, the current models have limited abilities to memorize the sequence of previous states during planning and to solve search problems in complicated environments (graphs). Both limitations lead the large language models incapable of adversarial planning. Overall, large language models can play significant roles in planning, especially few-shot planning, and their significance can be improved over time.

%% file: tex/A-appendix.tex
\newpage
\appendix

\section{Appendix}

\subsection{Planning Domain Examples}

\begin{lstlisting}[float=!htb, caption={Define a problem and a set of actions for the chess game in PDDL.}, label={lst:chess}, language=PDDL]
(define (domain chess)
  (:requirements :strips :typing :equality :conditional-effects)
  (:types piece square )
  (:predicates
    (clear ?s - square) 
    (at ?p - piece ?s - square) 
    (captured ?p - piece)
  )
  (:action move
    :parameters (?p - piece ?from - square ?to - square)
    :precondition (and
      (at ?p ?from) (clear ?to)
    )
    :effect (and
      (not (at ?p ?from)) (at ?p ?to) (clear ?from) (not (clear ?to))
    )
  )
  (:action capture
    :parameters (?p1 - piece ?p2 - piece ?from - square ?to - square)
    :precondition (and
      (at ?p1 ?from) (at ?p2 ?to)
    )
    :effect (and
      (not (at ?p1 ?from)) (at ?p1 ?to) (not (at ?p2 ?to))
      (captured ?p2) (clear ?from)
    )
  )
)

(define (problem chess-problem)
  (:domain chess)
  (:objects
    wp1 wp2 wp3 wp4 wp5 wp6 wp7 wp8 - piece
    bp1 bp2 bp3 bp4 bp5 bp6 bp7 bp8 - piece
    a1 a2 a3 a4 a5 a6 a7 a8
    ...
    h1 h2 h3 h4 h5 h6 h7 h8 - square
  )
  (:init
    (at wp1 a2) (at wp2 b2) (at wp3 c2) (at wp4 d2) (at wp5 e2) 
    (at wp6 f2) (at wp7 g2) (at wp8 h2)
    (at bp1 a7) (at bp2 b7) (at bp3 c7) (at bp4 d7) (at bp5 e7) 
    (at bp6 f7) (at bp7 g7) (at bp8 h7)
    (clear a3) (clear a4) (clear a5) (clear a6) (clear a8)
    ...
    (clear h3) (clear h4) (clear h5) (clear h6) (clear h8)
  )
  (:goal (and (at wp1 a7) ))
)

\end{lstlisting}